\documentclass{article}
\PassOptionsToPackage{table}{xcolor}
\usepackage[T1]{fontenc}
\usepackage{iclr2027_conference,times}
\renewcommand{\encodingdefault}{OT1}
\usepackage{amsmath,amsfonts,bm}

\def\eqref#1{equation~\ref{#1}}
\def\1{\bm{1}}

\DeclareMathAlphabet{\mathsfit}{\encodingdefault}{\sfdefault}{m}{sl}
\SetMathAlphabet{\mathsfit}{bold}{\encodingdefault}{\sfdefault}{bx}{n}

\usepackage{amsmath}
\usepackage{booktabs}
\usepackage{graphicx}
\usepackage{multirow}
\usepackage{array}
\usepackage[table]{xcolor}
\usepackage{tabularx}
\usepackage{microtype}
\usepackage{placeins}
\usepackage{float}
\usepackage{needspace}
\usepackage{hyperref}
\usepackage{url}
\newcommand{\ScaleAuditRows}{%
Description & 0.08368 & 0.05498 & 37.0\% & 13/24 & 0.991 \\
Matched full & 0.09129 & 0.05614 & 41.1\% & 11/24 & 1.011 \\
Within-task shuffle & 0.08413 & 0.05802 & 33.9\% & 10/24 & 1.063 \\
Cross-task shuffle & 0.09219 & 0.05747 & 40.3\% & 11/24 & 1.043 \\
Numeric only & 0.09041 & 0.05490 & 41.8\% & 11/24 & 1.007 \\
Prose only & 0.08791 & 0.05492 & 40.1\% & 10/24 & 1.037 \\
}
\newcommand{\OpenMLClassificationRows}{%
cmc & 23 & 1,473 & Public \\
steel-plates-fault & 146817 & 1,941 & Public \\
analcatdata\_dmft & 3560 & 797 & Public \\
first-order-theorem-proving & 9985 & 6,118 & Public \\
pc4 & 3902 & 1,458 & Public \\
credit-approval & 29 & 690 & Public \\
blood-transfusion-service-center & 10101 & 748 & Public \\
sick & 3021 & 3,772 & Public \\
phoneme & 9952 & 5,404 & Public \\
diabetes & 37 & 768 & Public \\
churn & 167141 & 5,000 & public \\
adult & 7592 & 48,842 & Public \\
}
\newcommand{\OpenMLRegressionRows}{%
health\_insurance & 361269 & 22,272 & GPL (\textgreater{}= 2) \\
cps88wages & 361261 & 28,155 & Public \\
student\_performance\_por & 361619 & 649 & CC BY 4.0 \\
socmob & 361264 & 1,156 & Non-commercial research \\
space\_ga & 361623 & 3,107 & Public \\
red\_wine & 361250 & 1,599 & CC BY 4.0 \\
abalone & 361234 & 4,177 & CC BY 4.0 \\
california\_housing & 361255 & 20,640 & Public \\
brazilian\_houses & 361267 & 10,692 & CC 0: Public Domain \\
miami\_housing & 361260 & 13,932 & CC0: Public Domain \\
kings\_county & 361266 & 21,613 & CC 0: Public Domain \\
fifa & 361272 & 19,178 & CC0: Public Domain \\
}
\newcommand{\OpenMLFrozenContrastRows}{%
Description minus full point MAE & $-0.007414$ & [$-0.028044$, $+0.002505$] & $-0.000472$ \\
Within shuffle minus full point MAE & $-0.007313$ & [$-0.037487$, $+0.006536$] & $+0.000964$ \\
Description minus full repeat drift & $-0.002963$ & [$-0.012119$, $+0.002542$] & $+0.000753$ \\
Description minus within repeat drift & $-0.002144$ & [$-0.011138$, $+0.002896$] & $+0.001391$ \\
}

\definecolor{auditblue}{HTML}{5271AE}
\definecolor{auditred}{HTML}{D85B59}
\definecolor{tablehead}{HTML}{DBE7F5}
\definecolor{tablestripe}{HTML}{F3F6FA}
\newcolumntype{Y}{>{\raggedright\arraybackslash}X}
\newcolumntype{R}{>{\raggedleft\arraybackslash}X}
\newcommand{\tablestyle}{%
  \setlength{\tabcolsep}{4pt}%
  \renewcommand{\arraystretch}{1.06}%
  \rowcolors{2}{white}{tablestripe}%
}
\newcommand{\code}[1]{\texttt{#1}}

\newcommand{\IS}{\mathrm{IS}_{80}}

\hypersetup{hidelinks,pdftitle={Predictive Credit: Measuring What Scientific Explanations Add to Experimental Forecasts}}
\title{Predictive Credit: Measuring What Scientific\\
Explanations Add to Experimental Forecasts}

\iclrfinalcopy
\author{%
\makebox[\dimexpr\textwidth-2\tabcolsep\relax][c]{%
\begin{tabular}{c}
Jingjie Ning\textsuperscript{1}\thanks{Corresponding author}\quad
Xueqi Li\textsuperscript{1}\quad
Yibo Kong\textsuperscript{1}\quad
Dongting Li\textsuperscript{2}\\[4pt]
{\normalfont\small \textsuperscript{1}Carnegie Mellon University\quad
\textsuperscript{2}Tsinghua University}\\[2pt]
{\normalfont\footnotesize
\{\href{mailto:jening@cs.cmu.edu}{jening},\,
\href{mailto:xueqil@cs.cmu.edu}{xueqil},\,
\href{mailto:yibok@cs.cmu.edu}{yibok}\}@cs.cmu.edu\quad
\href{mailto:ldt25@mails.tsinghua.edu.cn}{ldt25@mails.tsinghua.edu.cn}}
\end{tabular}}%
}
\hypersetup{pdfauthor={Jingjie Ning, Xueqi Li, Yibo Kong, Dongting Li}}

\def\arxivpreprint{}

\begin{document}
\maketitle
\pagestyle{plain}
\thispagestyle{plain}
\begin{abstract}
Research agents explain planned experiments. We measure predictive credit
with paired forecasts sharing an intervention, forecaster, and outcome while
varying description, matched explanation, and donor context. Five checks track
commitment, delivery, predictive gain, alignment, and known-signal uptake.
Across 336 prospective states in controlled learning, 12 Tox21 endpoints,
and 24 OpenML tasks, v5's frozen credit decision was inconclusive. Tox21's
preregistered ROC AUC interval-score harm test was unmet ($D-M=-.0026$,
95 percent interval [$-.0174$, .0104]); OpenML's joint formation,
point-equivalence, and repeatability rule was unmet. Matched point-accuracy
gains over description remained unconfirmed, and Tox21/OpenML seed-donor
intervals spanned zero. Under requested DeepSeek V4 Pro, matched and donor
cards reduced secondary Tox21 drift by 64.5 and 59.1 percent. A DeepSeek V4
Flash replay raised matched point MAE from .01823 to .02020 and missed
matched-donor interval-score equivalence. OpenML full-card assignment widened
nominal 80 percent intervals by 21 percent, with 49.3 percent coverage versus
51.4 percent for description and content in 66/144 cards. Direct-text Flash
delivered all 144 notes without detectable matched point-accuracy gain. A
researcher-authored mechanism positive control lowered point MAE by 2.60
percentage points versus description. The protocol measures predictive credit
for research-agent benchmarks and scientific forecasting; natural-explanation
credit remained unconfirmed at the tested donor resolutions.
\end{abstract}

\newcommand{\paperteaser}[1]{%
\begin{figure}[#1]
\centering
\includegraphics[width=\linewidth]{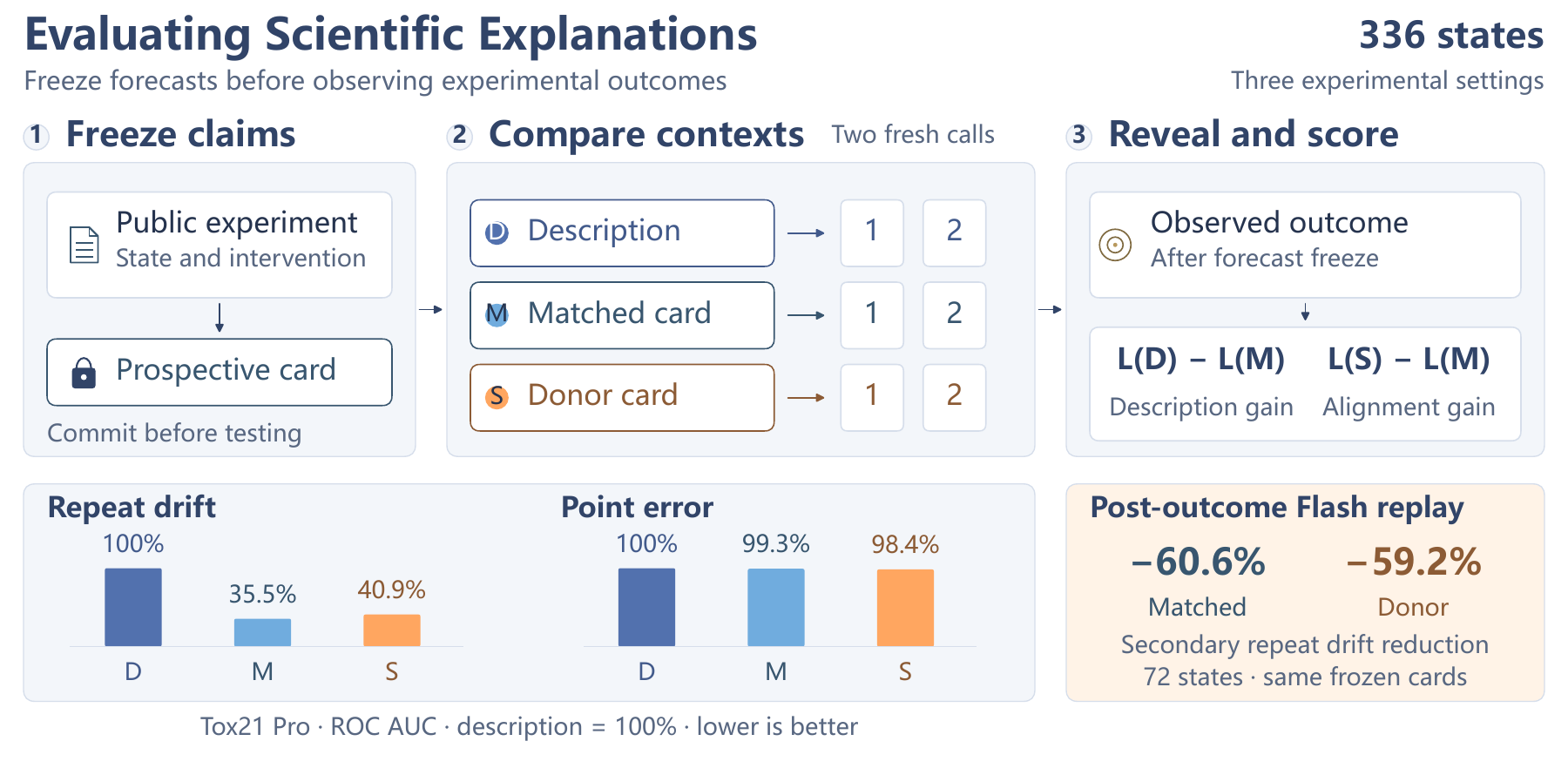}
\caption{\textbf{From a scientific claim to predictive credit.} Prospective
studies collect two fresh forecasts per context before outcomes are opened.
$L(D)$, $L(M)$, and $L(S)$ are observed losses whose context means form the
$R_c$ values in Eq.~\ref{eq:gains}. The Tox21 Pro bars show secondary repeat
drift and point MAE relative to description; its frozen primary criterion is
ROC AUC interval score. The post-outcome Flash replay uses the same 72 states
and cards.}
\label{fig:teaser}
\end{figure}
}
\ifdefined\arxivpreprint\paperteaser{H}\else\paperteaser{!h}\fi

\section{Introduction}

A scientific explanation earns practical value by helping anticipate the
consequences of an intervention, a planned experimental change.
An explanation states why that change should affect the outcome. For example,
a rationale for stronger regularization can predict its effect on
generalization or the training gap. These predictions connect scientific
reasoning to experimental resource allocation and give autonomous agents'
research proposals a common evaluation target.

Automated research systems connect idea generation, implementation, measurements,
and reporting \citep{lu2024ai,ning2026specialists}. Molecular workflows test
selected changes on held-out targets \citep{ning2026molecular}. These systems produce both an
experimental result and an account of why a proposed change should work.
Final performance evaluates the search outcome. Evaluating the accompanying
explanation requires asking what it contributes to predicting that outcome.
This question gives research-agent benchmarks a way to score experimental
rationales alongside achieved results \citep{bragg2025astabench}.

Scientific forecasting studies already predict empirical AI and neuroscience
results \citep{wen2025predicting,luo2025brainbench}. Simulatability methods
evaluate explanations by the model behavior they help an observer predict
\citep{hase2020leakage,chen2024simulatability,mayne2026positive}.
We connect these traditions by measuring \emph{predictive credit}, the gain in
forecasting executed outcomes from assigning an agent's explanation context.
The public state, comprising the available data and measurements, and the
planned intervention form the description baseline.
A donor explanation comes from another state under a frozen assignment.
It tests whether gains depend on matching the current case.
Both comparisons use the same
experimental outcome, forecaster, and loss.

The empirical challenge is that explanation-dependent behavior has several
observable forms. A prediction card organizes an explanation into explicit,
testable fields. It can make a claim explicit, bring repeated forecasts
closer together, change their accuracy, or widen their uncertainty intervals.
Our 336-state prospective program and follow-up controls measure these
responses together. Its frozen studies left predictive credit unconfirmed.
The Tox21 primary ROC AUC interval-score harm rule was unmet; structured
cards reduced secondary repeat drift in the prospective study and a Flash
replay. A component crossover tests numerical targets and mechanism text.
A direct-text OpenML pipeline measures full-note delivery and forecast quality.
A known-mechanism positive control improves accuracy while drift rises.

The paper makes three contributions. First, the paired design evaluates the
predictive value and local alignment of scientific explanations. Second, the
prospective studies and component controls separate commitment, repeatability,
and outcome information. Third, direct-text forecasts test delivered-note
value, while exact-signal calibration and a known-mechanism positive control
measure two forms of forecaster sensitivity.
Frozen records support benchmark reanalysis, scientific forecasting,
explanation evaluation, and methods for cost-aware experiment selection.

\section{Related work}

\paragraph{Forecasting scientific outcomes.}
\citet{wen2025predicting} evaluate pairwise predictions of empirical AI research
outcomes. BrainBench tests predictions of neuroscience findings from study
descriptions \citep{luo2025brainbench}.
\citet{mule2026forecast} train models to compare research ideas using their
benchmark outcomes. Research preference models select experiments using plans,
code, and optional pilot runs \citep{foster2026preference}.
CUSP evaluates the feasibility, mechanisms, solutions,
and timing of scientific advances under temporal knowledge constraints
\citep{wu2026cusp}. These studies establish scientific forecasting as an
evaluation target and motivate outcome-aware idea selection. Our paired
comparisons measure each rationale's gain for a fixed state and intervention.

\paragraph{The predictive value of explanations.}
Leakage-adjusted simulatability evaluates how explanations help an observer
predict model outputs while accounting for answer leakage
\citep{hase2020leakage}. Counterfactual simulatability extends evaluation to
related inputs \citep{chen2024simulatability}.
\citet{mayne2026positive} report gains from self-explanations and compare
explanations exchanged across models.
\citet{karvonen2026chive} test whether activation-based information improves
predictions of model behavior under counterfactual prompt edits. We evaluate
research-agent explanations through forecasts of numerical changes in external
experiments, with matched and donor cards, repeated forecasts, and uncertainty.
Broader faithfulness tests examine the relationship
between explanations and model decisions
\citep{turpin2023language,atanasova2023faithfulness,madsen2024selfexplanations};
\citet{parcalabescu2024measuring} distinguish this goal from output consistency.

\paragraph{Content attribution and repeated trials.}
Interventions on reasoning traces measure the influence of intermediate
text on generated answers \citep{lanham2023measuring}. Revision or Re-Solving
separates recomputation, structural scaffolding, and draft content
\citep{ning2026revision}. Same Agent, Different Answers compares corpus-induced
changes with ordinary repeat variability \citep{ning2026churn}. We repeat forecasts of
one fixed experiment and report both drift and predictive error.

\paragraph{Agent evaluation and research workflows.}
AstaBench evaluates scientific research tasks \citep{bragg2025astabench}.
Scientific-agent trace analysis examines evidence uptake and belief revision
{\fontencoding{T1}\selectfont\citep{riosgarcia2026scientists}}. EvoSCM commits causal hypotheses to
falsifiable predictions before experimental feedback \citep{zhao2026evoscm}.
Our paired evaluation measures the gain from a supplied explanation for fixed
executed interventions alongside task-level and trace-level assessments.

\paragraph{Measurement and uncertainty.}
Construct-validity work separates observed measurements from the concepts
they are intended to represent \citep{cronbach1955construct,borsboom2004concept}.
Language-model calibration relates confidence to correctness
\citep{kadavath2022language}, self-consistency can improve task answers
\citep{wang2023selfconsistency}, and semantic entropy estimates uncertainty
through variation in meaning across generations \citep{farquhar2024semantic}.
We measure forecast drift, point error, and proper interval score under
supplied-card interventions \citep{gneiting2007strictly}.

\section{Evaluating scientific explanations}
\label{sec:audit}

\subsection{Predictive value and local alignment}

Let $x_i$ be the public state of experiment $i$, $a_i$ its planned intervention,
$h_i$ a prospective explanation, and $y_i$ the realized change in an outcome
metric. A fixed forecaster $f$ receives one of three contexts
\begin{equation}
 D_i=(x_i,a_i),\qquad
 M_i=(x_i,a_i,h_i),\qquad
 S_i=(x_i,a_i,h_{\pi(i)}).
 \label{eq:contexts}
\end{equation}
The frozen mapping $\pi$ selects another state's explanation. \emph{Matched}
denotes the current state's explanation; \emph{donor} denotes the assigned
one. Stored \emph{shuffled} and \emph{wrong-seed} explanation labels mean
donor; \emph{focal} and \emph{aligned} mean matched. Calibration's other-seed
outcome is a numeric signal. Tox21 and OpenML swap seeds within the same
task, budget, and intervention cell. V5 pairs different interventions.
Post-outcome checks found
matching effect signs in 27/36 Tox21 and 54/72 OpenML seed pairs. Tox21's
median seed gap was .01362 ROC AUC against point MAE near .02.

In the source studies, $x_i$ contains public training summaries and parent-only
development results. Tox21 shows parent validation metrics; OpenML shows
parent validation loss, skill, and a prediction hash; v5 shows parent metrics
and learning curves. Tox21 and OpenML generators see the assigned action,
while v5 proposes from a public catalog. Every forecaster sees the parent state,
chosen action, and assigned note. Measured child-validation results,
operability-gate outputs, and held-out outcomes remain outside the prompts.
Thus $y_i$ is the held-out child-minus-parent effect forecast before child
measurements are supplied.

For a loss $\ell$, write $R_c=\mathbb{E}[\ell(f(c_i),y_i)]$. The two
predictive contrasts are
\begin{equation}
 G_{\mathrm{description}}=R_D-R_M,\qquad
 G_{\mathrm{alignment}}=R_S-R_M.
 \label{eq:gains}
\end{equation}
Positive values favor the matched context. The first contrast measures its
forecast gain over the public description. The second measures its advantage
over the study's assigned donor. Both use paired outcomes and a common
forecast interface. Delivery records identify evidence-backed content in the
assigned forecast inputs.

Predictive credit is relative to the forecaster, target, loss, and task
distribution. Joint positive gains, interpreted with the delivery records,
support state-specific credit at the tested donor resolution. The design extends
predictive-usefulness evaluation to executed experiments.

\begin{figure}[t]
\centering
\includegraphics[width=\linewidth]{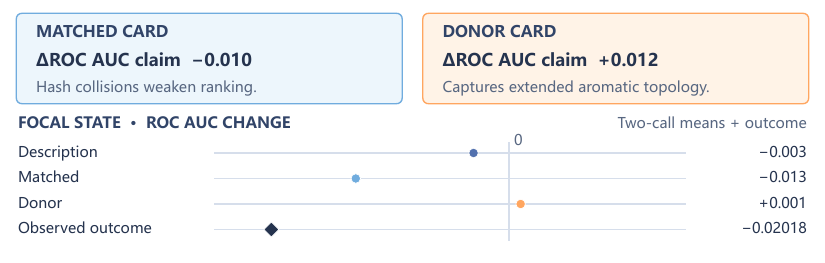}
\caption{\textbf{One Tox21 pair links claims and executed outcomes.}
The lexicographically first endpoint-budget cell is NR-AhR at 40\% labels;
the lower seed is the focal state. It expands the molecular fingerprint radius
from 2 to 3. Marks compare two-call means with its outcome. Selection used
frozen identities before outcomes. Across 72 states, matched/donor point MAE
was .02041/.02023.}
\label{fig:worked-example}
\end{figure}

\subsection{Commitment, agreement, and prediction}

\paragraph{Commitment.}
Commitment is the set of testable predictions stated before the outcome.
The formation and delivery analyses count six prediction-card slots. They
are an explicit target direction, a quantitative target point or interval,
a named intermediate observable, an expected benefit regime, a falsification
criterion, and a counterfactual. A complete card supplies all six slots.
Free-text extraction requires an exact supporting span for each slot.
Tox21 mechanism prose is screened separately in the component crossover.
Completeness counts these six slots.

\paragraph{Agreement.}
Two independent calls produce point forecasts $\hat y_{ic1}$ and
$\hat y_{ic2}$. Their repeat drift is
\begin{equation}
 A_{ic}=|\hat y_{ic1}-\hat y_{ic2}|.
 \label{eq:drift}
\end{equation}
Drift measures variation in repeated outputs. A shared numerical center can
reduce $A_{ic}$ while retaining a common error against $y_i$. Jointly reporting
drift and outcome loss distinguishes output coordination from predictive gain.

\paragraph{Prediction.}
Every call returns a point forecast and a central 80 percent interval
$[L,U]$. We report point MAE, direction accuracy, interval coverage, width,
and the proper interval score
\begin{equation}
 \IS(L,U;y)=(U-L)+10(L-y)\mathbf{1}[y<L]
                  +10(y-U)\mathbf{1}[y>U].
 \label{eq:score}
\end{equation}
The score rewards narrow intervals and penalizes missed outcomes
\citep{gneiting2007strictly}. Constant-zero and constant-direction forecasts
make the benefit of model inference visible against simple task priors.

\subsection{Five checks for predictive credit}

The five checks are prospective \emph{commitment}, evidence-backed
\emph{delivery}, paired predictive \emph{value} against description and simple
priors, donor \emph{alignment}, and \emph{sensitivity} to known signals.
All assigned calls remain in
intention-to-treat (ITT) analysis. An exhausted or invalid response receives
a deterministic fallback and reduces its stratum's integrity rate.
The integrity rate is the fraction of assigned calls with valid responses.
ITT measures the full pipeline, including these failures. Content-level analysis
also measures delivery, the presence of evidence-backed explanation fields in
the forecast input. Delivered-card comparisons describe the selected cases
where that content arrived.

Task-aware aggregation accompanies all five checks. Repeated seeds and label
budgets share a task, so task is the top inference unit in the external studies.
Scale diagnostics report constant-baseline error, taskwise ratios, and the
influence of leaving out each task. Equivalence means that an estimated
difference is small enough to fall within a prespecified practical margin.
A confidence interval entirely inside that margin
supports the corresponding equivalence statement. An interval extending across
the margin records the remaining range of plausible effects.

\section{Experimental design}
\label{sec:design}

\subsection{Three experimental settings}

We evaluate computational interventions spanning synthetic learning
regimes, real molecular assay-activity targets, and heterogeneous tabular
prediction tasks. The interventions modify features, objectives,
regularization, or training budget under fixed modeling families.
The three prospective studies contain 120, 72, and 144 states, respectively.
Tox21 and OpenML follow-up checks reuse these states; the mechanism positive control adds 40
derived states. Table~\ref{tab:studies} places selected frozen source-study
contrasts in the main text.

\begin{table}[H]
\caption{\textbf{Prospective paired loss gains.} Every row follows the
$D-M$ or $S-M$ direction in Eq.~\ref{eq:gains}; positive favors matched.
Two-sided intervals are 95\% for v5 and Tox21 and 90\% for OpenML. V5 donors
come from different interventions; Tox21 and OpenML donors swap seeds.}
\label{tab:studies}
\centering
\small
\tablestyle
\begin{tabularx}{\linewidth}{>{\raggedright\arraybackslash}p{.98in}Yrr}
\toprule
\rowcolor{tablehead}
\textbf{Study} & \textbf{Loss and signed contrast} & \textbf{Estimate} & \textbf{Interval} \\
\midrule
Controlled v5 & Accuracy point MAE, $D-M$ (pp) & $+.071$ & [$-.102$, +.252] \\
Controlled v5 & Log-loss point MAE, $D-M$ & $+.0363$ & [$-.0061$, +.0807] \\
Controlled v5 & Log-loss point MAE, $S-M$ & $+.0508$ & [+.0022, +.1076] \\
Controlled v5 & Accuracy interval score, $D-M$ & $-3.8436$ & [$-7.9180$, $-.5468$] \\
Controlled v5 & Log-loss interval score, $D-M$ & $-2.0878$ & [$-4.2911$, $-.2750$] \\
Tox21 & ROC AUC interval score, $D-M$ & $-.002628$ & [$-.017392$, +.010375] \\
Tox21 & ROC AUC interval score, $S-M$ & $-.002856$ & [$-.016750$, +.011428] \\
OpenML & Point MAE, $D-M$ & $-.00741$ & [$-.02804$, +.00251] \\
OpenML & Point MAE, $S-M$ & $-.00731$ & [$-.03749$, +.00654] \\
\bottomrule
\end{tabularx}
\end{table}

Tox21's frozen harm gate required $M-D$ and $M-S$ to reach .005 with positive
one-sided lower bounds; the gate returned \code{confirmation\_no\_go}.

\paragraph{Controlled v5.}
The 120 states split evenly between ordinary proposals and six-field card
elicitation. Ordinary proposals used quote-bound extraction. Each state
received two forecasts per context. Thirty ran preselected follow-ups; their
four-class effect-change predictions were correct in 11/30, matching simple
majority baselines.

\paragraph{Tox21 anchor.}
Twelve assay endpoints, three training-label fractions, and two seeds produced 72 states
\citep{wu2018moleculenet}. Molecular features feed a converged logistic model.
A cyclic assignment gave both seeds in each endpoint-budget cell the same
intervention. Each state produced one card and six forecasts. The donor swap
preserved endpoint, budget, action, and parent recipe. The primary outcome was
the interval score for ROC AUC change.

\paragraph{OpenML v2.1.}
A metadata-based hash lottery selected 12 classification tasks from
OpenML-CC18 and 12 regression tasks from OpenML-CTR23, with distinct source
families \citep{vanschoren2014openml,bischl2021openml,fischer2023ctr23}.
Three nested training-label budgets and two seeds produced 144 states. A LightGBM
pipeline received one of six assigned changes. Spontaneous and elicited notes
shared free-text output and condition-blind extraction. Each state received
two forecasts under description, matched, numeric, prose, within-task donor,
and cross-task donor contexts.
OpenML numeric-only retained extracted direction and point-or-interval slots,
counting either as nonempty; prose-only retained the other four slots. Tox21's
target-number component contains three target intervals and a benefit
probability. Classification and regression targets used log-loss and MSE skill
change, with skill $1-\mathrm{loss}/\mathrm{loss}_{null}$. Grouped splits and task
IDs appear in Appendix~\ref{app:lineage}.

Tool-free Claude CLI sessions accessed DeepSeek's Anthropic-compatible
endpoint. Source cards, forecasts, and calibration requested
\code{deepseek-v4-pro}; follow-up forecasts requested \code{deepseek-v4-flash}.
Both requested routes belong to the DeepSeek V4 family. Returned wrappers recorded usage
without a resolved model ID. V5 used high effort; other calls used low effort.
The 3,708 prospective slots froze before private outcomes; first responses,
requested routes, and data identities remain in the event ledgers.

\subsection{Targeted model replay and channel calibration}

The 432-call Flash replay reuses 72 Tox21 states, cards, and donors under a
pre-call frozen plan. The 1,440-call calibration supplies five known-signal
contexts across 144 OpenML states. Task-scaled MAE divides each absolute error
by the larger of its task's mean absolute outcome change and .01, averaging
six states per task and then 24 tasks. Component and raw-note follow-ups reuse
source states. The mechanism positive control adds 40 paired states.
Positive-control outcomes were computed before prompt freeze and withheld
from the forecaster.

\Needspace{7\baselineskip}
\subsection{Evidence status and statistical analysis}

The three source studies froze records before outcomes. Their rules tested
predictive value (v5), interval-score harm (Tox21), and a joint formation,
point-equivalence, and repeatability criterion (OpenML). V5 returned
\code{inconclusive}; Tox21 and OpenML returned \code{confirmation\_no\_go}.
Cross-study analyses and controls used existing outcomes under separately
frozen call plans; Appendix~\ref{app:lineage} records the gates.

Source losses average calls before state aggregation; the OpenML ensemble
sensitivity averages forecasts before scoring. Tox21 bootstraps endpoints,
budget cells, and seeds; OpenML stratifies by task kind and resamples tasks.
The 12 endpoints and 24 tasks are the external inference units. The post-outcome
crossover froze its fixed-number mechanism contrast before Flash calls; other
arm contrasts are exploratory and unadjusted for multiplicity.

\section{Separating repeatability from predictive gain}
\label{sec:agreement}

Tox21's frozen primary ROC AUC interval-score harm rule returned
\code{confirmation\_no\_go} (Table~\ref{tab:studies}). As a secondary result,
Pro repeat drift fell from .01004 under description to .00357 with matched
cards and .00411 with seed-level donors. The reductions were 64.5 and 59.1
percent; point MAE was .02056, .02041, and .02023, respectively.

\begin{table}[!htb]
\caption{\textbf{Tox21 repeatability and prediction.} Pro and Flash receive
the same cards. Drift is the absolute difference between two calls.
Point MAE and interval score average the two call losses.}
\label{tab:replay}
\centering
\small
\tablestyle
\begin{tabularx}{\linewidth}{YYRRRR}
\toprule
\rowcolor{tablehead}
\textbf{Predictor} & \textbf{Context} & \textbf{ROC drift} & \textbf{Reduction} & \shortstack{\bfseries Point\\\bfseries MAE} & \shortstack{\bfseries Interval\\\bfseries score} \\
\midrule
Pro & Description & .01004 & reference & .02056 & .08972 \\
Pro & Matched & .00357 & 64.5\% & .02041 & .09235 \\
Pro & Donor & .00411 & 59.1\% & .02023 & .08949 \\
\midrule
Flash & Description & .00704 & reference & .01823 & .08022 \\
Flash & Matched & .00278 & 60.6\% & .02020 & .09160 \\
Flash & Donor & .00288 & 59.2\% & .02013 & .08833 \\
\bottomrule
\end{tabularx}
\end{table}

The 432-call Flash replay reduced drift by 60.6 and 59.2 percent with matched
and donor cards. Matched-card point MAE rose from .01823 under description to
.02020; donor-card MAE was .02013. The frozen joint status was
\code{replication\_not\_supported} because interval-score equivalence exceeded
its $\pm .010$ margin (Table~\ref{tab:flash-equivalence}).

The 1,152-call Flash component crossover reused the same states, cards, and
full-card prompts. Its description baseline was .00427, versus .00704 in the
Flash replay. Crossover full-card drift was .00319, a reduction of .00108
with interval [$-.00053$, .00283], versus .00426 in the replay. The
numbers-only arm was one of seven card-versus-description contrasts; its
exploratory unadjusted drift estimate was .00181 with interval [.00005, .00383].
At fixed numbers, donor-minus-matched screened-mechanism point MAE was
$-.00047$ with interval [$-.00228$, .00104]. The two Flash runs show different
full-card drift magnitudes on the same stimuli. All seven card contexts had
point MAE above description's .01738 (range .01797 to .01873).

OpenML Pro mean drift was .01004 under description and .01300 with matched
cards. The description-minus-matched difference was $-.00296$ with 90 percent
interval [$-.01212$, .00254]; 10 percent trimmed drift was .00673 and .00610.
The rounded .01004 description values in Tox21 and OpenML use ROC AUC-change
and skill-change units, respectively. Direct-text Flash mean drift was .00488
and .00962 under description and matched notes.

\FloatBarrier
\section{Commitment and uncertainty}
\label{sec:proxies}

\begin{figure}[!htb]
\centering
\includegraphics[width=\linewidth]{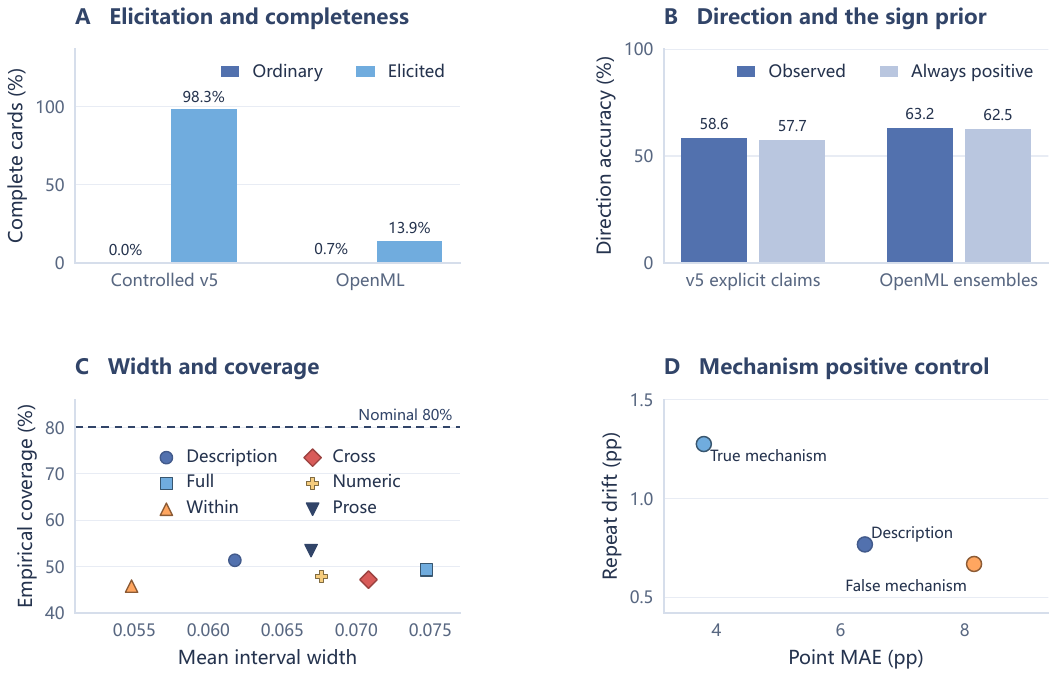}
\caption{\textbf{Different observables identify different behaviors.}
Formation varies across study and interface. Direction accuracy closely follows
the sign prior. OpenML coverage measures the placement of stated uncertainty.
The positive control compares point error and drift.}
\label{fig:proxy}
\end{figure}

\subsection{Elicitation changes commitment rates}

Controlled v5 produced complete cards in 0 of 60 ordinary proposals and
59 of 60 elicited proposals. The direct card schema created a strong
completion response. Under OpenML's common free-text and extraction path,
completeness was 1 of 144 versus 20 of 144, a difference of 13.2 percentage
points with a one-sided 95 percent lower bound of 6.9 points. The two
experiments compare complete interface designs, combining environment,
effort, schemas, and extraction. Card completeness therefore measures the
commitments elicited by each deployed interface.

Prediction supplies a separate criterion. In v5, direction accuracy conditional
on a claim was 59.6 percent under ordinary prompting and 57.6 percent under
elicitation. Elicited central-80-percent card intervals covered 33 of 59
outcomes, or 55.9 percent. Matched log-loss point MAE was .1905 versus .2268
for description and .2413 for the different-intervention donor. The donor
minus matched gain was +.0508 with interval [+.0022, +.1076]. Accuracy and
log-loss interval scores increased from 19.30 to 23.14 and from 1.41 to
3.50. Table~\ref{tab:studies} reports the frozen paired contrasts.

\subsection{A sign prior explains much of direction accuracy}

Among v5's 111 explicit directions, 109 predicted improvement. Their accuracy
was 65/111, or 58.6 percent; always predicting improvement scored 64/111,
or 57.7 percent. All 52 ordinary-prompt directions said improvement, exactly
reproducing that baseline's 59.6 percent accuracy. OpenML description-only
accuracy was 63.19 percent against a constant-positive baseline of 62.5 percent.
On the 86 states with $|y|\geq .01$, the two-call direction rule and the
constant-positive baseline both scored 68.6 percent. This latter comparison is
a post hoc movement sensitivity; the full threshold sweep appears in the
appendix. Constant baselines quantify the contribution of local direction forecasts.

\subsection{Wider intervals coexist with undercoverage}

OpenML forecast intervals cover 45.8 to 53.5 percent of outcomes across the
six contexts, below the nominal 80 percent target. In intention-to-treat
analysis, assigning the full-card context increased mean interval width from
.06179 to .07474, about 21 percent. Seventy-eight of 144 assigned cards
were empty scaffolds. The paired width change was .01294 with a task-stratified
95 percent interval of [.00141, .03306]. Coverage is 49.3 percent with full
cards and 51.4 percent with description. The coverage difference is
$-2.08$ percentage points with interval [$-6.25$, $+2.08$]; the interval-score
difference is .02567 with interval [$-.01512$, .08214]. These intervals
establish a width increase. Width and coverage intervals use 95 percent
inference; the frozen OpenML point-equivalence rule uses 90 percent intervals.

On the post hoc $|y|\geq .01$ subset, coverage is 24.4 to 30.8 percent.
These conditional rates characterize the difficulty of larger realized effects.
V5 shows a related width-without-coverage pattern. Accuracy width grows from 4.63 to
9.37 percentage points as coverage changes from 59.2 to 61.3 percent.
Log-loss width grows from .295 to 2.796 as coverage changes from 48.8 to
51.3 percent. The audit links expressed uncertainty to both its empirical
coverage and its proper score.

\FloatBarrier
\Needspace{5\baselineskip}
\section{Delivery, scale, and detectable information}
\label{sec:interpretation}

\begin{figure}[!htbp]
\centering
\includegraphics[width=\linewidth]{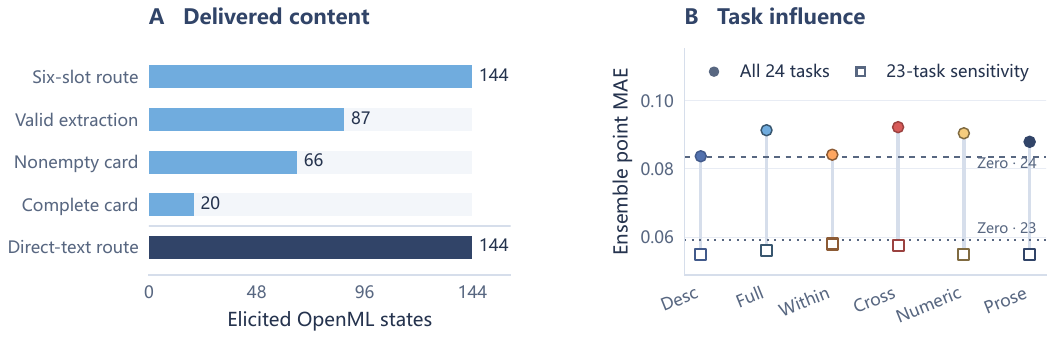}
\caption{\textbf{The analysis population determines the interpretation.}
The delivery panel compares six-slot extraction and direct-text delivery.
The scale panel shows pooled ensemble MAE with and without
\code{brazilian\_houses}, alongside constant-zero baselines.}
\label{fig:delivery}
\end{figure}

\subsection{Delivery defines the content comparison}

OpenML evaluates two forecast pipelines with distinct representations and
predictors. The prospective Pro pipeline starts from 144 elicited notes with
text. Condition-blind six-slot extraction produced 87 valid responses, 66
nonempty cards, 20 complete cards, and 78 empty scaffolds. Among 72 donor
pairs, 13 had content on both sides and one had quantitative claims on both.

The post-outcome Flash pipeline supplied all 144 Pro-authored source notes as
raw text. All 864 Flash forecasts were valid.
Description, matched, and donor point MAEs were .08257, .08311, and .08197.
Task-balanced scaled matched gains against description and donor were $-.0906$
and .0064, with 95 percent intervals [$-.2687$, .0403] and [$-.0964$, .0898].
The raw-MAE donor-minus-matched gain was $-.00114$; task scaling changes the
relative weight of task magnitudes and reverses this point-estimate sign.
The direct-text Flash results show no detectable matched point-accuracy
advantage within the direct-text pipeline.
Its descriptive interval scores were .67866, .63622, and .63872 for
description, matched, and donor notes; corresponding coverages were 50.0,
58.3, and 56.9 percent.

\subsection{Task scale determines influence on the aggregate}

The six \code{brazilian\_houses} states in the 144-state population account
for 37.0 percent of description error and 41.1 percent of matched-full error.
Overall description ensemble MAE is .08368 against .08341 for a zero forecast.
Omitting this task gives .05498 for description and .05614 for matched full,
against a .05910 zero baseline. The skill scale
$1-\mathrm{MSE}/\mathrm{MSE}_{null}$ amplifies concentrated errors.

Taskwise ratios and leave-one-task-out influence expose concentrated error.
On the original scale, description-minus-full MAE was $-.00741$ with a
90 percent interval of [$-.02804$, .00251]; within-shuffle-minus-full was
$-.00731$ with interval [$-.03749$, .00654]. Both intervals extend below $-.01$.

\subsection{Positive controls for known information}

The researcher-authored mechanism positive control paired 40 linear and
nonlinear tasks under one quadratic intervention. The public state omitted
the train-only quadratic residual diagnostic. The true note disclosed the
latent functional form, which strongly predicts whether quadratic features
help, while withholding measured target outcomes. All 240 Flash forecasts
were valid. Point MAE
was 3.795 with the true note, 6.391 with description, and 8.151 with the false
note. The true note gained 2.596 percentage points over description with a
95 percent pair-bootstrap interval of [2.314, 2.880]. Its 4.355-point
advantage over the false note had interval [4.048, 4.654]. This controlled
clue tests uptake of strong known mechanism information.
The executed quadratic intervention raised held-out accuracy by 11.44 points
in nonlinear tasks and changed it by $-1.06$ points in paired linear tasks.

The separate exact-signal calibration gave 144/144 exact-signal wins and a
point-value Spearman correlation of .99996. Its frozen status was
\code{assay\_sensitive}.

\section{Applications in research evaluation}
\label{sec:reuse}

The five checks yielded higher elicited-card completeness, with OpenML's .132
gain below its frozen .50 formation threshold; content in 66/144 OpenML cards;
unconfirmed joint predictive credit; seed-donor intervals spanning zero in
Tox21 and OpenML; and an \code{assay\_sensitive} exact-signal control. The
protocol scores explanations on executed experiments. A cost-aware utility
rule can rank proposals using forecasters that pass paired predictive checks.

\section{Conclusion}

We measure predictive credit through paired forecasts of fixed interventions.
Across 336 prospective states, description, matched, and donor contexts share
each outcome and forecaster.

The frozen v5 decision was inconclusive; Tox21 and OpenML returned
\code{confirmation\_no\_go}. Tox21's primary ROC AUC interval-score harm rule
was unmet. Structured cards reduced secondary drift in Tox21, with variable
full-card magnitudes across Flash runs. OpenML full-card assignment widened
intervals by 21 percent with 78 empty cards; direct-text Flash delivered all
144 notes without a detectable matched point-accuracy gain.

A researcher-authored mechanism positive control and separate exact-signal
calibration demonstrate uptake of supplied information. The protocol separates
commitment, delivery, repeatability, and accuracy for research-agent benchmarks.
Forecast-based ranking requires demonstrated predictive gain and a task-specific
cost-utility rule.

\subsubsection*{Ethics statement}
The experiments use existing benchmark datasets and computational model
changes. Molecular targets measure assay activity. The task manifest records
data sources and supplied license strings.

\subsubsection*{Reproducibility statement}
\ifdefined\arxivpreprint
Appendix~\ref{app:lineage} details the study designs, task IDs, donor rules,
and analyses. We retain first responses, source snapshots, frozen reports,
and exact OpenML split assignments for audit.
\else
The anonymous supplement contains protocols, source snapshots, first
responses, donor maps, ledgers, analysis inputs, frozen reports, and an
offline verifier. It records OpenML task IDs and exact split assignments.
Appendix~\ref{app:lineage} details the study designs.
\fi

\subsubsection*{AI use statement}
Language models assisted literature retrieval, implementation, experiment
orchestration, analysis, manuscript drafting and editing, and figures.
The experimental calls retain their requested DeepSeek model routes, raw
responses, source snapshots, and frozen analysis records.

\begingroup
\renewcommand{\encodingdefault}{T1}
\normalfont
\bibliography{references}
\endgroup
\bibliographystyle{iclr2027_conference}

\clearpage
\appendix

\FloatBarrier
\Needspace{10\baselineskip}
\section{Study design and evidence timing}
\label{app:lineage}

\FloatBarrier
\Needspace{8\baselineskip}
\subsection{Evidence status}

Controlled v5, Tox21 anchor, and OpenML v2.1 are prospective source studies.
The cross-study measurement synthesis and assay calibration are post hoc
analyses using known source outcomes. The 1,440 calibration calls followed
a separate frozen protocol and gate. The Tox21 Flash replay followed its
own post-outcome frozen protocol and gate. Three Flash follow-up controls
tested components, direct raw-note delivery, and a mechanism positive control.
Table~\ref{tab:lineage} records each study's timing,
population, and role.

\begin{table}[!htbp]
\caption{Evidence lineage and population accounting}
\label{tab:lineage}
\centering
\scriptsize
\tablestyle
\begin{tabularx}{\linewidth}{YrrrrY}
\toprule
\rowcolor{tablehead}
\textbf{Study} & \textbf{Tasks} & \textbf{States} & \textbf{Pre-outcome} & \textbf{Post-outcome} & \textbf{Role} \\
\midrule
Controlled v5 & 1 environment & 120 & 900 & 0 & Prospective source study \\
Tox21 anchor & 12 endpoints & 72 & 504 & 0 & Prospective source study \\
OpenML v2.1 & 24 tasks & 144 & 2,304 & 144 & Prospective source study \\
Assay calibration & 24 tasks & 144 reused & 0 & 1,440 & Post-outcome sensitivity \\
Flash replay & 12 endpoints & 72 reused & 0 & 432 & Post-outcome model-version replay \\
Component crossover & 12 endpoints & 72 reused & 0 & 1,152 & Post-outcome component test \\
Raw-note replay & 24 tasks & 144 reused & 0 & 864 & Post-outcome direct-text pipeline \\
Mechanism positive control & 1 environment & 40 new & 0 & 240 & Known-signal uptake \\
\midrule
Unique prospective total & 3 settings & 336 & 3,708 &  &  \\
\bottomrule
\end{tabularx}
\end{table}

The 40 positive-control states are paired tasks outside the 336
prospective source states. Their deterministic outcomes were computed from a
fixed data-generating design before the public states, mechanism notes, and
Flash call plan were frozen. Every positive-control prompt withheld the target outcome.
The component and raw-note follow-ups reused source outcomes already known
when their prompts and analyses were frozen.

\FloatBarrier
\Needspace{8\baselineskip}
\subsection{Controlled v5 design}

The 120 states crossed six study-authored fast machine learning regimes and
frozen random seeds. Sixty states were assigned to spontaneous proposals and
sixty to explicit elicitation. Every state received two independent forecasts
in each of three contexts. The shuffled hypothesis came from another
intervention under a frozen derangement. The primary prediction targets were
accuracy change in percentage points and log-loss change. Thirty cases also
executed the follow-up counterfactual named before the primary outcome.

The 900 required units consisted of 120 proposals, 60 spontaneous-note
extractions, and 720 forecasts. A missing proposal left one extraction slot
unissued. All assigned units remained in the denominator.

\FloatBarrier
\Needspace{8\baselineskip}
\subsection{Tox21 anchor design}

The Tox21 state lattice was 12 endpoints by three label budgets by two seeds.
One action was assigned to every endpoint-budget cell using
$a=(\text{endpoint index}+\text{budget index})\bmod 6$. Both seeds in a cell
received the same action. The endpoint index follows the frozen Tox21 catalog
order beginning NR-AR, NR-AR-LBD, NR-AhR. NR-AhR at 40\% labels has endpoint
index 2 and budget index 0, assigning larger Morgan radius. Actions were balanced loss, physicochemical features,
larger Morgan radius, wider Morgan fingerprint, increased L2, and decreased L2.

Cards and predictors saw confirmation-development information only. The
confirmation-outcome labels stayed closed until one card and six predictor
artifacts per state had frozen. The shuffled control swapped the two seed cards
within the exact endpoint, budget, action, parent recipe, and stage.

For $\alpha=.20$, the interval score was

\begin{equation}
(U-L)+\frac{2}{\alpha}(L-y)\mathbf{1}[y<L]+\frac{2}{\alpha}(y-U)\mathbf{1}[y>U].
\end{equation}

The two calls were averaged within state. Seeds were averaged inside
endpoint-budget cells, budgets inside endpoints, and endpoints received equal
weight. The hierarchical bootstrap resampled endpoint, then budget cell, then
seed. The frozen harm claim required both matched-minus-control estimates to be
at least .005, both one-sided lower confidence limits (LCLs) and trimmed means to be positive,
joint positivity in at least 8 endpoints and 4 actions, positivity at every
budget, complete population, and integrity of at least .95 in every stratum.

\FloatBarrier
\Needspace{8\baselineskip}
\subsection{OpenML v2.1 design}

The confirmation tasks were selected by a metadata-only hash lottery from
OpenML-CC18 and OpenML-CTR23. Conservative source families were unique across
discovery and confirmation. Images, OCR, free text, chemistry, artificial data,
games, simulations, and tasks with explicit temporal or grouped dependence were
excluded. The selected confirmation tasks appear in
Tables~\ref{tab:openml-cls} and \ref{tab:openml-reg}.

Rows with identical encoded features form a group and stay in one partition.
The study-authored split assigns groups to training, validation, and private
outcomes in approximately 60/20/20 proportions. Classification uses class
strata; regression uses target-quantile strata, with a stable random fallback
for small populations. The recorded task identifiers specify the source data.
The frozen study records contain the exact split memberships and sampling seeds.

Each task used nested 25, 50, and 100 percent training-label budgets and two
frozen seeds. A cyclic assignment balanced six interventions across task-budget
combinations. They
were L1 regularization, stronger L2 regularization, feature subsampling, row
subsampling, 100 boosting rounds, and 400 boosting rounds. A zero-LLM public
operability gate fitted parent and child on public training data and verified
different configuration and validation-prediction hashes in all 144 states.
It used validation features without decoding their labels, computing child
loss or skill, or opening held-out outcomes. Gate hashes and Boolean results
stayed in separate artifacts and never entered note or predictor prompts.
The model prompts contained parent public-validation loss, skill, and
a prediction hash alongside the assigned action. They asked agents to forecast
the private held-out skill change before measured child validation results
were supplied.

The two note arms shared a minimal free-text output schema. A condition-blind,
quote-bound extractor mapped each note to six possible slots. The six predictor
contexts each received two independent calls. The within-task shuffle swapped
the two seeds in the exact task, budget, and action cell. Cross-task shuffles
preserved task kind, action, budget, and seed while changing task. Numeric-only
retained direction and point-or-interval slots; prose-only retained observable,
benefit regime, falsifier, and counterfactual slots. Nonempty component input
required at least one present retained slot. The Tox21 component crossover
used target-metric intervals and benefit probability for its numerical arm.

The v2.1 protocol retained exhausted two-attempt calls as terminal ITT
fallbacks. Each logical call allowed two attempts; fallbacks counted toward
the prespecified .95 integrity gate.

\begin{table}[!htbp]
\caption{OpenML classification confirmation tasks}
\label{tab:openml-cls}
\centering
\small
\tablestyle
\begin{tabularx}{\linewidth}{Yrrl}
\toprule
\rowcolor{tablehead}
\textbf{Dataset} & \textbf{OpenML task} & \textbf{Rows} & \textbf{Recorded license} \\
\midrule
\OpenMLClassificationRows
\bottomrule
\end{tabularx}
\end{table}

\begin{table}[!htbp]
\caption{OpenML regression confirmation tasks}
\label{tab:openml-reg}
\centering
\small
\tablestyle
\begin{tabularx}{\linewidth}{Yrrp{1.6in}}
\toprule
\rowcolor{tablehead}
\textbf{Dataset} & \textbf{OpenML task} & \textbf{Rows} & \textbf{Recorded license} \\
\midrule
\OpenMLRegressionRows
\bottomrule
\end{tabularx}
\end{table}

\FloatBarrier
\Needspace{8\baselineskip}
\subsection{Assay calibration design}

The calibration reused all 144 OpenML states after outcomes were known. Exact
signal supplied the true skill change. Noisy signal added a deterministic sign
times half the larger of task mean absolute outcome and .01. The shuffled signal
used the exact outcome from the other seed in the same task, budget, and action
cell. Call order, noise signs, donors, and values were frozen before the first
calibration call. Two independent calls were issued in each of five contexts.

The gate required exact signal to beat description in at least 75 percent of
states with a one-sided task-bootstrap lower bound above 65 percent. It also had
to beat wrong-seed signal in at least 70 percent with lower bound above 60
percent. Forecast and exact hint Spearman correlation had to be at least .80.
Every condition and call-index integrity stratum had to be at least .95.

\FloatBarrier
\Needspace{10\baselineskip}
\section{Controlled-study results}

\FloatBarrier
\Needspace{8\baselineskip}
\subsection{Formation and direction}

\begin{table}[!htbp]
\caption{Controlled v5 prediction-card formation}
\centering
\small
\tablestyle
\begin{tabularx}{\linewidth}{Yrr}
\toprule
\rowcolor{tablehead}
\textbf{Measure} & \textbf{Spontaneous} & \textbf{Elicited} \\
\midrule
Complete card & 0/60 (0.0\%) & 59/60 (98.3\%) \\
Quantitative magnitude & 1/60 (1.7\%) & 59/60 (98.3\%) \\
Explicit direction & 52/60 (86.7\%) & 59/60 (98.3\%) \\
Intermediate observable & 36/60 (60.0\%) & 59/60 (98.3\%) \\
Direction correct given claim & 31/52 (59.6\%) & 34/59 (57.6\%) \\
Central-80 interval covered & Insufficient intervals & 33/59 (55.9\%) \\
\bottomrule
\end{tabularx}
\end{table}

Of 111 explicit directions, 109 were improvement, one was decline, and one was
unchanged. Nine states had no explicit direction. Overall direction accuracy
was 65/111 (58.6 percent) while always predicting improvement scored 64/111
(57.7 percent). Search succeeded in 71/120 states (59.2 percent). The frozen
report's binary search-prediction correlation of .869 arises under nearly
constant direction predictions, leaving the intended separation unidentified.
Point forecasts correlated .4964 with actual outcomes.

\FloatBarrier
\Needspace{8\baselineskip}
\subsection{Forecasts and primary contrasts}

\begin{table}[!htbp]
\caption{Controlled v5 condition metrics}
\centering
\scriptsize
\tablestyle
\begin{tabularx}{\linewidth}{llRRRRR}
\toprule
\rowcolor{tablehead}
\textbf{Target} & \textbf{Context} & \textbf{Direction} & \shortstack{\bfseries Point\\\bfseries MAE} & \textbf{Coverage} & \textbf{Width} & \shortstack{\bfseries Interval\\\bfseries score} \\
\midrule
Accuracy & Description & .6250 & 3.0945 pp & .5917 & 4.6277 pp & 19.2976 \\
Accuracy & Matched hypothesis & .6458 & 3.0235 pp & .6125 & 9.3667 pp & 23.1412 \\
Accuracy & Shuffled hypothesis & .6583 & 3.0486 pp & .6167 & 6.1750 pp & 20.2275 \\
Log loss & Description & .7208 & .2268 & .4875 & .2948 & 1.4118 \\
Log loss & Matched hypothesis & .7333 & .1905 & .5125 & 2.7960 & 3.4996 \\
Log loss & Shuffled hypothesis & .7292 & .2413 & .5042 & 1.2183 & 2.0355 \\
\bottomrule
\end{tabularx}
\end{table}

\begin{table}[!htbp]
\caption{Controlled v5 frozen contrasts}
\centering
\small
\tablestyle
\begin{tabularx}{\linewidth}{Yrr}
\toprule
\rowcolor{tablehead}
\textbf{Contrast with positive meaning matched is better} & \textbf{Estimate} & \textbf{Paired 95\% interval} \\
\midrule
Description minus matched accuracy MAE & +0.071 pp & [$-0.102$, +0.252] \\
Description minus matched log-loss MAE & +0.0363 & [$-0.0061$, +0.0807] \\
Shuffled minus matched accuracy MAE & +0.025 pp & [$-0.186$, +0.240] \\
Shuffled minus matched log-loss MAE & +0.0508 & [+0.0022, +0.1076] \\
Description minus matched accuracy interval score & $-3.8436$ & [$-7.9180$, $-0.5468$] \\
Description minus matched log-loss interval score & $-2.0878$ & [$-4.2911$, $-0.2750$] \\
\bottomrule
\end{tabularx}
\end{table}

The two-call point drifts for description, matched, and shuffled were .8611,
.5587, and 1.0205 percentage points for accuracy. They were .1427, .0861, and
.1240 for log loss. Intermediate-observable direction accuracy was .645, .640,
and .621 for description, matched, and shuffled. Counterfactual transport
measures correct predictions of how the effect changes under the follow-up
intervention. Its accuracy was
11/30, or 36.7 percent, matching the always-stronger and always-same majority
baselines. The
agent predicted 14 weaker, 13 stronger, 3 reverse, and 0 same, while 11 actual
cases were same.

\FloatBarrier
\Needspace{8\baselineskip}
\subsection{Integrity and cost}

\begin{table}[!htbp]
\caption{Controlled v5 operational integrity}
\centering
\small
\tablestyle
\begin{tabularx}{\linewidth}{Yrrr}
\toprule
\rowcolor{tablehead}
\textbf{Stratum} & \textbf{Units} & \textbf{Response} & \textbf{Semantic} \\
\midrule
Proposal & 120 & 99.17\% & 98.33\% \\
Spontaneous extraction & 60 & 98.33\% & 98.33\% \\
Description forecast & 240 & 100.00\% & 100.00\% \\
Matched forecast & 240 & 98.33\% & 97.50\% \\
Shuffled forecast & 240 & 99.17\% & 99.17\% \\
\bottomrule
\end{tabularx}
\end{table}

There were 892 assistant responses, seven empty transport failures, and one
unissued extraction. All failures remained as deterministic fallbacks.
Successful-response usage reconstructs to USD 6.011690. The repository's
artifact inventory records token counts and alternative-price diagnostics.

\FloatBarrier
\Needspace{10\baselineskip}
\section{Tox21 results and heterogeneity}

\FloatBarrier
\Needspace{8\baselineskip}
\subsection{Frozen decision}

\begin{table}[!htbp]
\caption{Tox21 frozen central-80-percent ROC AUC interval-score contrasts}
\centering
\small
\tablestyle
\begin{tabularx}{\linewidth}{Yrrrr}
\toprule
\rowcolor{tablehead}
\textbf{Matched minus control} & \textbf{Estimate} & \shortstack{\bfseries One-sided\\\bfseries LCL} & \textbf{Two-sided 95\% interval} & \textbf{Trimmed} \\
\midrule
Description & +.002628 & $-.008521$ & [$-.010375$, +.017392] & $-.000281$ \\
Shuffled card & +.002856 & $-.008943$ & [$-.011428$, +.016750] & +.003041 \\
\bottomrule
\end{tabularx}
\end{table}

Positive values mean worse matched-card interval score. Both estimates were
below the frozen .005 minimum. Five of 12 endpoints and three of six actions
had both contrasts positive. The required counts were eight and four. Both
contrasts changed sign across label budgets.

\FloatBarrier
\Needspace{8\baselineskip}
\subsection{All condition metrics}

\begin{table}[!htbp]
\caption{Tox21 condition metrics over all 72 states}
\centering
\scriptsize
\tablestyle
\begin{tabularx}{\linewidth}{llRRRRR}
\toprule
\rowcolor{tablehead}
\textbf{Target} & \textbf{Context} & \shortstack{\bfseries Point\\\bfseries MAE} & \textbf{Coverage} & \textbf{Width} & \shortstack{\bfseries Interval\\\bfseries score} & \shortstack{\bfseries Repeat\\\bfseries drift} \\
\midrule
ROC AUC & Description & .02056 & .79861 & .06585 & .08972 & .01004 \\
ROC AUC & Matched & .02041 & .82639 & .06788 & .09235 & .00357 \\
ROC AUC & Shuffled & .02023 & .81250 & .06565 & .08949 & .00411 \\
Average precision & Description & .03349 & .74306 & .09091 & .18952 & .01535 \\
Average precision & Matched & .03238 & .75000 & .08717 & .17041 & .00585 \\
Average precision & Shuffled & .03375 & .73611 & .08474 & .17158 & .00608 \\
Log loss & Description & .03469 & .73611 & .06585 & .23516 & .01339 \\
Log loss & Matched & .03512 & .72917 & .06327 & .23796 & .00467 \\
Log loss & Shuffled & .03554 & .70833 & .06117 & .23646 & .00475 \\
\bottomrule
\end{tabularx}
\end{table}

\begin{table}[!htbp]
\caption{Tox21 prospective-card calibration}
\centering
\small
\tablestyle
\begin{tabularx}{\linewidth}{Yrrrr}
\toprule
\rowcolor{tablehead}
\textbf{Card target} & \shortstack{\bfseries Point\\\bfseries MAE} & \textbf{Coverage} & \textbf{Width} & \shortstack{\bfseries Interval\\\bfseries score} \\
\midrule
ROC AUC & .02129 & .79167 & .06876 & .09541 \\
Average precision & .03565 & .70833 & .09017 & .17749 \\
Log loss & .03484 & .69444 & .06057 & .23063 \\
\bottomrule
\end{tabularx}
\end{table}

\begin{table}[!htbp]
\caption{Tox21 secondary matched-card contrasts}
\centering
\scriptsize
\tablestyle
\begin{tabularx}{\linewidth}{lYYrr}
\toprule
\rowcolor{tablehead}
\textbf{Target} & \textbf{Quantity} & \textbf{Control} & \textbf{Estimate} & \textbf{Two-sided 95\% interval} \\
\midrule
ROC AUC & Interval-score harm & Description & +.002628 & [$-.010375$, +.017392] \\
ROC AUC & Interval-score harm & Shuffled & +.002856 & [$-.011428$, +.016750] \\
ROC AUC & Point-MAE harm & Description & $-.000148$ & [$-.003028$, +.002752] \\
ROC AUC & Point-MAE harm & Shuffled & +.000183 & [$-.002112$, +.002600] \\
Average precision & Interval-score harm & Description & $-.019108$ & [$-.053253$, +.010825] \\
Average precision & Interval-score harm & Shuffled & $-.001168$ & [$-.025866$, +.019123] \\
Average precision & Point-MAE harm & Description & $-.001119$ & [$-.006469$, +.003806] \\
Average precision & Point-MAE harm & Shuffled & $-.001377$ & [$-.005879$, +.002707] \\
Log loss & Interval-score harm & Description & +.002806 & [$-.039257$, +.043349] \\
Log loss & Interval-score harm & Shuffled & +.001504 & [$-.026402$, +.027990] \\
Log loss & Point-MAE harm & Description & +.000423 & [$-.004342$, +.004949] \\
Log loss & Point-MAE harm & Shuffled & $-.000425$ & [$-.003023$, +.002234] \\
\bottomrule
\end{tabularx}
\end{table}

\FloatBarrier
\Needspace{8\baselineskip}
\subsection{Endpoint, action, and budget diagnostics}

\begin{table}[!htbp]
\caption{Tox21 endpoint-specific interval-score harm}
\centering
\small
\tablestyle
\begin{tabularx}{\linewidth}{YrrYrr}
\toprule
\rowcolor{tablehead}
\textbf{Endpoint} & \textbf{vs desc} & \textbf{vs shuffle} & \textbf{Endpoint} & \textbf{vs desc} & \textbf{vs shuffle} \\
\midrule
NR-AR & +.03600 & +.02509 & NR-PPAR-GAMMA & +.00564 & +.01647 \\
NR-AR-LBD & $-.01768$ & $-.01468$ & SR-ARE & +.00310 & +.00658 \\
NR-AHR & $-.00414$ & $-.00461$ & SR-ATAD5 & +.00602 & $-.01412$ \\
NR-AROMATASE & +.00038 & +.00525 & SR-HSE & +.01993 & +.00022 \\
NR-ER & $-.00383$ & +.00250 & SR-MMP & $-.01231$ & +.02050 \\
NR-ER-LBD & $-.00183$ & +.00092 & SR-P53 & +.00026 & $-.00986$ \\
\bottomrule
\end{tabularx}
\end{table}

\begin{table}[!htbp]
\caption{Tox21 action-specific and budget-specific interval-score harm}
\centering
\small
\tablestyle
\begin{tabularx}{\linewidth}{YrrYrr}
\toprule
\rowcolor{tablehead}
\textbf{Action} & \textbf{vs desc} & \textbf{vs shuffle} & \textbf{Budget} & \textbf{vs desc} & \textbf{vs shuffle} \\
\midrule
Balanced loss & +.00991 & +.01053 & .40 & +.01169 & $-.00005$ \\
Physchem features & +.00322 & +.00683 & .70 & $-.00884$ & +.00539 \\
Larger radius & +.01275 & $-.00268$ & 1.00 & +.00503 & +.00323 \\
Wider fingerprint & $-.00319$ & $-.01142$ & & & \\
Increase L2 & $-.01699$ & $-.00233$ & & & \\
Decrease L2 & +.01006 & +.01621 & & & \\
\bottomrule
\end{tabularx}
\end{table}

All 72 cards, 432 forecasts, and 72 outcomes were present. All eight required
integrity strata equaled 1.0. Four first attempts had empty transport failures
and succeeded on their single frozen retry. Successful DeepSeek usage cost
USD 2.87118. The repository records retries and costs.

\FloatBarrier

\FloatBarrier
\Needspace{10\baselineskip}
\section{Flash replay and equivalence checks}

The replay issued exactly 432 new Flash forecasts over the 72 frozen Tox21
states. It reused the 72 Pro-generated cards and their frozen model outcomes. Every condition and call-index integrity stratum equaled 1.0. All 432
calls requested the explicit \code{deepseek-v4-flash} route. No transport retry,
semantic fallback, or terminal fallback occurred.

\begin{table}[!htbp]
\caption{Flash replay condition metrics over all 72 Tox21 states}
\centering
\scriptsize
\tablestyle
\begin{tabularx}{\linewidth}{llRRRRR}
\toprule
\rowcolor{tablehead}
\textbf{Metric} & \textbf{Condition} & \shortstack{\bfseries Point\\\bfseries MAE} & \textbf{Coverage} & \textbf{Width} & \shortstack{\bfseries Interval\\\bfseries score} & \shortstack{\bfseries Repeat\\\bfseries drift} \\
\midrule
ROC AUC & Description & .01823 & .75694 & .05458 & .08022 & .00704 \\
ROC AUC & Matched & .02020 & .79861 & .06559 & .09160 & .00278 \\
ROC AUC & Shuffled & .02013 & .80556 & .06436 & .08833 & .00288 \\
Average precision & Description & .02643 & .70139 & .06304 & .15783 & .00876 \\
Average precision & Matched & .03188 & .68750 & .07972 & .17211 & .00410 \\
Average precision & Shuffled & .03199 & .72917 & .07947 & .16633 & .00326 \\
Log loss & Description & .03301 & .65278 & .04038 & .22732 & .00676 \\
Log loss & Matched & .03516 & .70139 & .05701 & .23846 & .00315 \\
Log loss & Shuffled & .03570 & .68056 & .05408 & .23993 & .00312 \\
\bottomrule
\end{tabularx}
\end{table}

For the frozen ROC AUC rule, matched repeat drift fell by 60.55 percent and
shuffled repeat drift fell by 59.17 percent. The corresponding absolute
reductions were .004264 and .004167. Their one-sided 95 percent lower bounds
were .002153 and .001931. The matched minus shuffled drift estimate was
$-.000097$ with 95 percent interval [$-.001597$, .001444]. The matched minus
shuffled point-MAE estimate was .000074 with interval [$-.002419$, .002768].

The strict conjunction returned \code{replication\_not\_supported}. Eight of
nine gates passed. The sole failure was interval-score equivalence. Matched
minus shuffled interval score was .003268 with 95 percent interval
[$-.008096$, .014485], extending above the frozen $\pm .010$ margin.

Matched minus description interval score was .011379 with interval
[$-.003901$, .026635]. The replay reproduces the reduction in repeat drift
across Pro and Flash. Full predictive-score equivalence remains unresolved at
the fixed margin.

Reconstructed successful DeepSeek usage was USD 0.497115. The source
repository identifies the frozen report and predictor responses.

\FloatBarrier

\FloatBarrier
\Needspace{10\baselineskip}
\section{OpenML results and sensitivity analyses}

\FloatBarrier
\Needspace{8\baselineskip}
\subsection{Population and intervention outcomes}

The final population contained 144 route states, 288 notes, 288 extractions,
1,728 forecasts, 144 outcomes, and 144 narrators. This gave 2,304 prospective
and 2,448 total logical calls. Ninety of 144 intervention effects were positive
and 54 were negative. One hundred twenty-eight had absolute skill change at
least .001. Median absolute skill change was .0236264.

\FloatBarrier
\Needspace{8\baselineskip}
\subsection{Formation and delivery}

\begin{table}[!htbp]
\caption{OpenML six-slot extraction and completeness}
\centering
\scriptsize
\setlength{\tabcolsep}{3.5pt}
\tablestyle
\begin{tabularx}{\linewidth}{lRRRRRRr}
\toprule
\rowcolor{tablehead}
\textbf{Arm} & \textbf{Complete} & \textbf{Direction} & \shortstack{\bfseries Point or\\\bfseries interval} & \textbf{Observable} & \textbf{Benefit} & \textbf{Falsifier} & \textbf{Counterfactual} \\
\midrule
Spontaneous & 1 & 62 & 8 & 61 & 49 & 27 & 58 \\
Elicited & 20 & 60 & 31 & 45 & 45 & 46 & 47 \\
\bottomrule
\end{tabularx}
\end{table}

The equal-task complete-card difference was .131944. Its one-sided 95 percent
lower bound was .069444 and its 90 percent interval was [.069444, .201389].
Fifteen of 24 tasks were positive. Classification and regression effects were
.097222 and .166667. Budget effects were .166667, .166667, and .062500 for 25,
50, and 100 percent labels. The frozen requirements were effect at least .50,
lower bound above .40, and at least 18 positive tasks.

\begin{table}[!htbp]
\caption{OpenML treatment delivery}
\centering
\small
\tablestyle
\begin{tabularx}{\linewidth}{Yrr}
\toprule
\rowcolor{tablehead}
\textbf{Quantity} & \textbf{Spontaneous} & \textbf{Elicited} \\
\midrule
Assigned notes & 144 & 144 \\
Valid extractor response & 118 (81.94\%) & 87 (60.42\%) \\
Nonempty extracted card & 93 (64.58\%) & 66 (45.83\%) \\
All-empty extracted card & 51 (35.42\%) & 78 (54.17\%) \\
Complete extracted card & 1 (0.69\%) & 20 (13.89\%) \\
\bottomrule
\end{tabularx}
\end{table}

Matched full and both shuffled full-card conditions delivered nonempty content
in 66 states each. Numeric-only content was nonempty in 63 and prose-only
content in 54. Matched and within-shuffled cards were both nonempty in 26
states. Matched and cross-shuffled cards were both nonempty in 26 states. These
subsets are post hoc delivery sensitivities on selected populations.
The 63 numeric-only deliveries are the union of 60 direction slots and 31
point-or-interval slots, with 28 cards containing both. Thirty-two cards
supplied direction alone and three supplied a point or interval alone.
Nonempty therefore records retained-slot coverage; a quantitative magnitude
was present in 31 elicited cards. The four prose slots had a 54-card union.

\FloatBarrier
\Needspace{8\baselineskip}
\subsection{Condition-level forecasts}

\begin{table}[H]
\caption{OpenML call-level condition metrics under the frozen scoring convention}
\label{tab:openml-call-level}
\centering
\scriptsize
\tablestyle
\begin{tabularx}{\linewidth}{lRRRRRR}
\toprule
\rowcolor{tablehead}
\textbf{Context} & \shortstack{\bfseries Point\\\bfseries MAE} & \shortstack{\bfseries Trimmed\\\bfseries MAE} & \textbf{Direction} & \textbf{Coverage} & \textbf{Width} & \shortstack{\bfseries Interval\\\bfseries score} \\
\midrule
Description & .08428 & .04111 & .6319 & .5139 & .06179 & .67655 \\
Matched full & .09169 & .04316 & .6181 & .4931 & .07474 & .70222 \\
Within-task shuffle & .08438 & .04462 & .6042 & .4583 & .05481 & .69159 \\
Cross-task shuffle & .09302 & .04361 & .6111 & .4722 & .07082 & .72461 \\
Matched numeric & .09080 & .04198 & .6042 & .4792 & .06764 & .71371 \\
Matched prose & .08820 & .03976 & .6250 & .5347 & .06694 & .70371 \\
\bottomrule
\end{tabularx}
\end{table}

Under the frozen call-level convention in Table~\ref{tab:openml-call-level},
always predicting a positive effect scored .6250 direction accuracy.
Full-card minus description width was +.0129432 with task-stratified
95 percent interval [.0014148, .0330630]. Coverage difference was
$-.0208333$ with interval [$-.0625$, .0208333]. Interval-score difference
was +.0256672 with interval [$-.0151208$, .0821369]. Positive values mean
higher interval loss for the matched full card.

\begin{table}[H]
\caption{OpenML two-call ensemble sensitivity metrics}
\label{tab:openml-ensemble}
\centering
\scriptsize
\tablestyle
\begin{tabularx}{\linewidth}{lRRRRRR}
\toprule
\rowcolor{tablehead}
\textbf{Context} & \shortstack{\bfseries Point\\\bfseries MAE} & \textbf{Direction} & \textbf{Coverage} & \textbf{Width} & \shortstack{\bfseries Interval\\\bfseries score} & \textbf{Spearman} \\
\midrule
Description & .08368 & .6319 & .5347 & .06179 & .67181 & .33133 \\
Matched full & .09129 & .6181 & .5069 & .07474 & .69845 & .37397 \\
Within-task shuffle & .08413 & .6042 & .4792 & .05481 & .68929 & .27711 \\
Cross-task shuffle & .09219 & .6111 & .4722 & .07082 & .72003 & .30057 \\
Matched numeric & .09041 & .6042 & .4722 & .06764 & .70872 & .36873 \\
Matched prose & .08791 & .6250 & .5903 & .06694 & .70089 & .30498 \\
\bottomrule
\end{tabularx}
\end{table}

For the two-call ensembles in Table~\ref{tab:openml-ensemble}, the predict-zero
pooled MAE was .0834107. Matched and description interval scores were .69845
and .67181, respectively, a descriptive difference of .02664. The frozen
interval-score inference above uses individual calls.

\Needspace{8\baselineskip}
\subsection{Frozen point and repeatability contrasts}

\begin{table}[!htbp]
\caption{OpenML frozen point contrasts and repeatability}
\centering
\small
\tablestyle
\begin{tabularx}{\linewidth}{Yrrr}
\toprule
\rowcolor{tablehead}
\textbf{Quantity} & \textbf{Estimate} & \textbf{90\% interval} & \textbf{Paired 10\% trim} \\
\midrule
\OpenMLFrozenContrastRows
\bottomrule
\end{tabularx}
\end{table}

The description-minus-full point contrast was +.000655 for classification and
$-.015482$ for regression. Its budget means were +.003123, $-.004192$, and
$-.021172$. The within-minus-full contrast was +.003094 for classification and
$-.017720$ for regression. Its budget means were +.004833, +.001469, and
$-.028240$. The hierarchical intervals extended beyond the frozen
equivalence margin of $\pm .01$.

Mean repeat drifts for description, full, within shuffle, cross shuffle,
numeric, and prose were .01004, .01300, .01218, .01626, .00965, and .01835.
Their 10 percent trimmed values were .00673, .00610, .00568, .00696, .00552,
and .00629. The frozen rule required description-to-full and
description-to-within reductions of at least .002 and 30 percent with positive
lower bounds in both task kinds. It also required full and within drift to be
equivalent within $\pm .002$. These conditions failed.

\FloatBarrier
\Needspace{8\baselineskip}
\subsection{Scale and null-baseline audit}

\begin{table}[!htbp]
\caption{OpenML scale influence and taskwise zero baselines}
\centering
\scriptsize
\tablestyle
\begin{tabularx}{\linewidth}{lRRRRR}
\toprule
\rowcolor{tablehead}
\textbf{Context} & \textbf{Pooled MAE} & \shortstack{\bfseries Without\\\bfseries worst task} & \textbf{Worst share} & \textbf{Tasks beat zero} & \shortstack{\bfseries Median\\\bfseries task ratio} \\
\midrule
\ScaleAuditRows
\bottomrule
\end{tabularx}
\end{table}

The worst task was \code{ctr23\_confirmation\_regression\_09} for every
condition. It is the \code{brazilian\_houses} data set. The taskwise mean MAE
ratios to each task's zero baseline were 1.145, 1.576, 1.422, 1.722, 1.370, and
1.247 for description, full, within, cross, numeric, and prose. The corresponding
medians were .991, 1.011, 1.063, 1.043, 1.007, and 1.037. The gap between means
and medians shows the influence of high-error tasks and motivates reporting
taskwise performance alongside pooled error.

\FloatBarrier
\Needspace{8\baselineskip}
\subsection{Post hoc movement sweep}

\begin{figure}[!htbp]
\centering
\includegraphics[width=\linewidth]{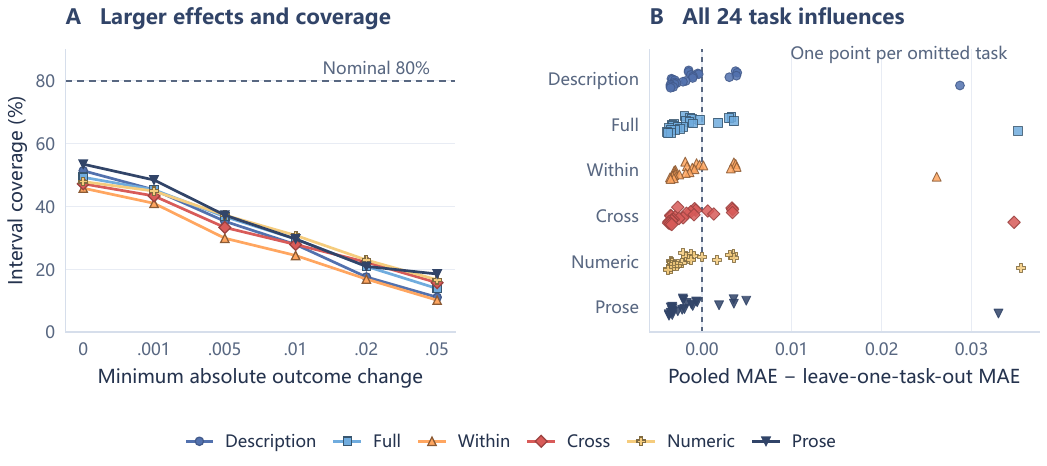}
\caption{\textbf{Outcome size and task composition shape forecast evaluation.}
Left, call-level interval coverage on nested subsets selected by realized
outcome size. These are post hoc descriptive comparisons. Right, each point
shows the change in pooled ensemble MAE after omitting one of the 24 tasks.
Positive values indicate lower error after omission. Colors identify each
forecast context.}
\label{fig:sensitivity}
\end{figure}

Figure~\ref{fig:sensitivity} shows interval coverage at increasing minimum
absolute outcome changes and each task's influence on pooled error.
The thresholded populations contain 144, 128, 102, 86, 74, and 54 states.
At threshold .01, description
direction accuracy was 69.77 percent at call level and 68.60 percent under the
two-call state convention. The constant-positive state rule was also 68.60
percent. Coverage decreased as realized movement grew.

\FloatBarrier
\Needspace{8\baselineskip}
\subsection{Post hoc treatment-delivery sensitivities}

\begin{table}[!htbp]
\caption{OpenML delivery sensitivities under two-call ensembles}
\centering
\scriptsize
\tablestyle
\begin{tabularx}{\linewidth}{Yrrrrr}
\toprule
\rowcolor{tablehead}
\textbf{Subset and contexts} & \textbf{$n$} & \textbf{Left MAE} & \textbf{Right MAE} & \shortstack{\bfseries Left\\\bfseries coverage} & \shortstack{\bfseries Right\\\bfseries coverage} \\
\midrule
Matched nonempty versus description & 66 & .08574 & .06893 & .5303 & .6061 \\
Matched and within both nonempty & 26 & .05820 & .06631 & .5769 & .5385 \\
Matched and cross both nonempty & 26 & .10697 & .09544 & .6538 & .6538 \\
Matched empty scaffold versus description & 78 & .09599 & .09616 & .4872 & .4744 \\
\bottomrule
\end{tabularx}
\end{table}

For matched nonempty versus description, interval scores were .55905 and
.50611. For matched and within both nonempty, they were .46112 and .53691. For
matched and cross both nonempty, they were .56555 and .51192. For the empty
scaffold subset, they were .81641 and .81203. These post hoc comparisons
describe performance conditional on treatment delivery. Estimates are unstable
across the selected subsets; the paired content-bearing comparisons each
contain 26 states.

\FloatBarrier
\Needspace{8\baselineskip}
\subsection{Donor resolution and paired cases}
\label{sec:donor-resolution}

The frozen within-cell seed pairs provide a post-outcome measure of how much
realized effects can differ when explanations are exchanged. Across 36 Tox21 endpoint-budget
pairs, the median absolute difference in realized ROC AUC change was .01362
(interquartile range [.00519, .02155]); 27 pairs had effects with the same sign.
Across 72 OpenML task-budget pairs, the corresponding skill-change difference
was .01504 ([.00364, .04742]); 54 pairs had effects with the same sign. Among
the OpenML elicited cards, 13 pairs contained evidence-backed content on both
sides, and one pair contained a point or interval claim on both sides. These
counts characterize the task and content resolution available to the
within-task alignment control.

We selected cases using cell order and card availability before inspecting
their outcomes. The lexicographically first Tox21 endpoint-budget cell is
\texttt{tox21\_nr\_ahr} at 40\% label
budget; both seeds increased Morgan radius from 2 to 3. The focal card
predicted a $-.010$ ROC AUC change from fingerprint collisions, while the donor
card predicted $+.012$ from extended aromatic topology. The realized changes
were $-.02018$ and $-.00969$. For the focal state, mean forecasts over two
calls were $-.003$ with the description, $-.013$ with the matched card, and
$+.001$ with the donor card. The matched explanation produced a more accurate
forecast in this illustrative cell.

\Needspace{4\baselineskip}
The sole OpenML pair with numeric slots on
both sides concerned L2 regularization at 50\% budget in the first CTR23
regression task. Its two cards predicted
overlapping positive ranges of $[+.005,+.020]$ and $[+.005,+.030]$; realized
changes were $-.00082$ and $+.00358$. The focal state's description, matched,
and donor forecasts were $+.0065$, $+.0120$, and $+.0125$. This second case
shows how the same evaluation captures low discrimination when paired cards
carry similar numerical claims.

\FloatBarrier
\Needspace{8\baselineskip}
\subsection{Integrity and usage}

Notes, forecasts, outcomes, and narrators had integrity 1.0. Spontaneous
extraction integrity was 118/144, or 81.94 percent. Elicited extraction integrity
was 87/144, or 60.42 percent. Classification and regression rates were 84.72 and
79.17 percent for spontaneous extraction, and 59.72 and 61.11 percent for
elicited extraction. All six required pooled and task-kind extractor strata
failed the .95 gate. Eighty-three exhausted calls stayed as ITT fallbacks.

The event ledger records every logical call and its DeepSeek V4 Pro route.
Successful usage reconstructs to USD 9.59019. The repository retains usage
details and cost bounds.

\FloatBarrier

\FloatBarrier
\Needspace{10\baselineskip}
\section{Known-signal calibration}

\begin{table}[!htbp]
\caption{Post-outcome positive-control condition metrics over all 144 states}
\centering
\scriptsize
\tablestyle
\begin{tabularx}{\linewidth}{lRRRRRR}
\toprule
\rowcolor{tablehead}
\textbf{Context} & \shortstack{\bfseries Point\\\bfseries MAE} & \textbf{Scaled MAE} & \textbf{Coverage} & \textbf{Width} & \textbf{Score} & \textbf{Spearman} \\
\midrule
Description replay & .092282 & .91949 & .5625 & .09079 & .71638 & .27505 \\
Empty card & .087885 & .91408 & .5069 & .07074 & .70540 & .17406 \\
Exact outcome signal & .0000156 & .00125 & .9931 & .01372 & .01372 & .99996 \\
Noisy outcome signal & .042405 & .49923 & .8333 & .08736 & .08741 & .79942 \\
Wrong-seed outcome signal & .046468 & .63636 & .1181 & .01311 & .44667 & .73789 \\
\bottomrule
\end{tabularx}
\end{table}

Exact signal beat description and wrong-seed signal in all 144 states.
The corresponding one-sided task-bootstrap lower bounds were 1.0, and
condition-by-call integrity was at least .99306. The frozen calibration
returned \code{assay\_sensitive}. Its complete call ledger and operational
records remain in the anonymous supplement.

\FloatBarrier
\section{Flash follow-up controls}
\label{app:flash-followups}

\subsection{Tox21 numerical and mechanism components}

The component crossover uses the 72 frozen Tox21 public states and
prospective Pro-authored cards. Its eight Flash contexts are description,
original full card, focal target numbers, screened focal mechanism text,
their matched combination, a donor mechanism at fixed focal numbers,
donor numbers at fixed focal mechanism, and a fully reconstructed donor.
Every state receives two independent forecasts per context. Screening
removes sentences that directly forecast target metrics while retaining
mechanistic process statements and public parent measurements. The 72
screening records and all 1,152 planned prompts were fixed before predictor
calls. The primary point-MAE contrast compares matched and donor mechanisms
while holding target numbers fixed. The reverse swap holds mechanism text
fixed and tests the donor target numbers. Endpoint, budget cell, and seed
form the bootstrap hierarchy.

\begin{table}[H]
\caption{Tox21 Flash component crossover over all 72 states}
\centering
\small
\tablestyle
\begin{tabularx}{\linewidth}{Yrrrr}
\toprule
\rowcolor{tablehead}
\textbf{Context} & \textbf{ROC drift} & \textbf{Point MAE} & \textbf{Interval score} & \textbf{Coverage} \\
\midrule
Description & .00427 & .01738 & .08332 & .722 \\
Original full card & .00319 & .01813 & .08674 & .806 \\
Focal target numbers & .00246 & .01820 & .08655 & .819 \\
Screened mechanism & .00378 & .01820 & .08884 & .729 \\
Focal numbers and mechanism & .00317 & .01845 & .09045 & .812 \\
Focal numbers, donor mechanism & .00311 & .01797 & .08923 & .819 \\
Donor numbers, focal mechanism & .00256 & .01855 & .08534 & .826 \\
Donor numbers and mechanism & .00310 & .01873 & .08510 & .868 \\
\bottomrule
\end{tabularx}
\end{table}

\begin{table}[H]
\caption{Flash practical-equivalence checks. Signed 95 percent intervals are
compared with the frozen margins; a pass requires full interval inclusion.}
\label{tab:flash-equivalence}
\centering
\scriptsize
\tablestyle
\begin{tabularx}{\linewidth}{Ylrrr}
\toprule
\rowcolor{tablehead}
\textbf{Signed comparison} & \textbf{Metric} & \textbf{Margin} & \textbf{95\% interval} & \textbf{Inside} \\
\midrule
Replay matched minus wrong seed & Drift & $\pm .002$ & [$-.001597$, .001444] & Yes \\
Replay matched minus wrong seed & Point MAE & $\pm .005$ & [$-.002419$, .002768] & Yes \\
Replay matched minus wrong seed & Interval score & $\pm .010$ & [$-.008096$, .014485] & No \\
Crossover full minus reconstructed & Drift & $\pm .002$ & [$-.001597$, .001542] & Yes \\
Crossover full minus reconstructed & Point MAE & $\pm .005$ & [$-.001699$, .000859] & Yes \\
\bottomrule
\end{tabularx}
\end{table}

All 1,152 Flash calls were valid. Target numbers alone reduced drift relative
to description by .00181, with a 95 percent endpoint-bootstrap interval of
[.00005, .00383]. The complete card reduced drift by .00108 with interval
[$-.00053$, .00283]. At fixed focal numbers, donor-minus-matched mechanism
point MAE was $-.00047$ with interval [$-.00228$, .00104]. At fixed focal
mechanism, the donor-number contrast was .00010 with interval
[$-.00183$, .00181]. Original-full minus reconstructed drift was +.00003.
Original-full minus reconstructed point MAE was $-.00032$, consistent with
the .01813 and .01845 means in the preceding table. Their intervals lie
inside the frozen $\pm .002$ drift and $\pm .005$ point-MAE margins in
Table~\ref{tab:flash-equivalence}. The contemporaneous
fixed-number estimate measures incremental state alignment over seed-level
donor prose while focal numbers remain shared. The numbers-only drift contrast
is exploratory and unadjusted across eight contexts. Full-card drift reduction
was .00108 here and .00426 in the 432-call replay using the same prompts and
states.

\subsection{Direct-text OpenML forecasts}

This post-outcome replay supplied all 144 frozen Pro-authored elicited OpenML notes directly
to a Flash forecaster. The public states, assigned interventions, and within-task
seed donors remained fixed. Each state received two fresh forecasts under
description, matched raw note, and donor raw note. All 864 assigned calls returned
valid first responses. Both matched and donor contexts contained the source
note text in all 144 states. The prospective pipeline used a Pro predictor
with extracted six-slot cards; this replay used Flash with raw note text.

\begin{table}[H]
\caption{OpenML raw-note Flash replay over the complete 144-state population}
\centering
\small
\tablestyle
\begin{tabularx}{\linewidth}{Yrrrr}
\toprule
\rowcolor{tablehead}
\textbf{Context} & \textbf{Point MAE} & \textbf{Interval score} & \textbf{Coverage} & \textbf{Repeat drift} \\
\midrule
Description & .08257 & .67866 & .5000 & .00488 \\
Matched raw note & .08311 & .63622 & .5833 & .00962 \\
Within-task donor raw note & .08197 & .63872 & .5694 & .00510 \\
\bottomrule
\end{tabularx}
\end{table}

Task-balanced scaled point-MAE gain for matched versus description was
$-.09059$ with a 95 percent task-bootstrap interval of [$-.26870$, .04033].
Matched versus donor gain was .00638 with interval [$-.09641$, .08983]. The
corresponding raw-MAE gains were $-.00054$ and $-.00114$. Direct delivery
identifies the predictive behavior of content-bearing notes across the full
assigned population. The point estimates were close, and both gain intervals
spanned zero.

\subsection{Known-mechanism positive control}

Twenty seeds from the controlled learning environment each produced a paired
linear and nonlinear label-generating task. Each task received the same assigned
quadratic-feature intervention. The pair shared its seed, sample sizes, noise
level, and parent recipe. The public state omitted the train-only quadratic
residual diagnostic. A researcher-authored mechanism note stated the latent
functional form and supplied strong directional information about the quadratic
intervention. The donor note
described the paired alternative functional form. Forty states received two
fresh Flash forecasts in each of the description, matched-note, and donor-note
contexts, yielding 240 valid first responses.

\begin{table}[H]
\caption{Flash forecasts under true and paired false mechanism clues}
\centering
\small
\tablestyle
\begin{tabularx}{\linewidth}{Yrrrr}
\toprule
\rowcolor{tablehead}
\textbf{Context} & \textbf{Point MAE} & \textbf{Interval score} & \textbf{Coverage} & \textbf{Repeat drift} \\
\midrule
Description & 6.391 & 49.435 & .488 & .768 \\
True mechanism & 3.795 & 18.210 & .663 & 1.275 \\
Paired false mechanism & 8.151 & 58.361 & .413 & .669 \\
\bottomrule
\end{tabularx}
\end{table}

The true mechanism reduced point MAE by 4.355 percentage points relative to
the paired false mechanism. The 95 percent same-seed-pair bootstrap interval
was [4.048, 4.654], with a one-sided lower bound of 4.094. Relative to
description, the reduction was 2.596 points with interval [2.314, 2.880].
The description contrast measures gain over the public state alone. The
paired-false contrast also includes the cost of a misleading mechanism clue.
The executed quadratic intervention raised held-out accuracy by 11.440 points
on average in the nonlinear tasks and changed it by $-1.055$ points in the
paired linear tasks. Mechanism-conditioned forecasts moved toward these
outcomes while retaining visible uncertainty.

\end{document}